%% file: main.tex
\documentclass{article}
\usepackage{amsmath}
\usepackage{amsfonts}
\usepackage{floatrow}
\usepackage{microtype}
\usepackage{graphicx}
\usepackage{subfigure}
\usepackage{booktabs} 
\usepackage{multirow}
\usepackage{lipsum}
\usepackage{hyperref}

\newcommand{\gray}[1]{\textcolor{gray}{}} 

\newcommand{\tae}{OPQL}

\usepackage[accepted]{icml2021}

\icmltitlerunning{Reasoning Over Virtual Knowledge Bases With Open Predicate Relations}

\begin{document}

\twocolumn[
\icmltitle{Reasoning Over Virtual Knowledge Bases \\ With Open Predicate Relations}



\icmlsetsymbol{equal}{*}

\begin{icmlauthorlist}
\icmlauthor{Haitian Sun *}{cmu}
\icmlauthor{Pat Verga}{goog}
\icmlauthor{Bhuwan Dhingra}{goog}
\icmlauthor{Ruslan Salakhutdinov}{cmu}
\icmlauthor{William W. Cohen}{goog}

\end{icmlauthorlist}

\icmlaffiliation{goog}{Google Research}
\icmlaffiliation{cmu}{Carnegie Mellon University}

\icmlcorrespondingauthor{Haitian Sun (* Most work done at Google Research)}{haitians@cs.cmu.edu}
\icmlcorrespondingauthor{Pat Verga}{patverga@google.com}
\icmlcorrespondingauthor{Bhuwan Dhingra}{bdhingra@google.com}
\icmlcorrespondingauthor{Ruslan Salakhutdinov}{rsalakhu@cs.cmu.edu}
\icmlcorrespondingauthor{William W. Cohen}{wcohen@google.com}

\icmlkeywords{Machine Learning, ICML}

\vskip 0.3in
]



\printAffiliationsAndNotice{} 

\begin{abstract}
We present the Open Predicate Query Language (OPQL); a method for constructing a virtual KB (VKB) trained entirely from text. Large Knowledge Bases (KBs) are indispensable for a wide-range of industry applications such as question answering and recommendation. Typically, KBs encode world knowledge in a structured, readily accessible form derived from laborious human annotation efforts. Unfortunately, while they are extremely high precision, KBs are inevitably highly incomplete and automated methods for enriching them are far too inaccurate. Instead, OPQL constructs a VKB by encoding and indexing a set of relation mentions in a way that naturally enables reasoning and can be trained without any structured supervision. We demonstrate that OPQL outperforms prior VKB methods on two different KB reasoning tasks and, additionally, can be used as an external memory integrated into a language model (OPQL-LM) leading to improvements on two open-domain question answering tasks.
\end{abstract}

\input{intro.tex}

\input{model.tex}
\input{reasoning.tex}
\input{memlm.tex}
\input{related_work.tex}

\section{Conclusion \label{sec:conclusion}}
We proposed \tae{} that can construct a virtual knowledge base from a text corpus without any supervision from existing KB. The pretrained \tae{} can effectively solve relational following task, achieving the state-of-the-art performance on two multi-hop relational following datasets. The improvement is more significant if evaluated on queries with relations not seen at training time. \tae{} can be injected into a language model to answer open-domain questions. It outperforms several large pretrained language models on two benchmark open-domain QA datasets.

\section{Acknowledgement}
This work was supported in part by NSF IIS1763562.

\input{output.bbl}
\bibliographystyle{icml2021}

\newpage \ \newpage
\input{appendix}

\end{document}

%% file: intro.tex
\section{Introduction \label{sec:intro}}

Large knowledge bases (KBs) structure information around triples of entities and relation types that describe the relationships between the subject and object entities, for example, [\textit{Charles Darwin}, author of, \textit{On the Origin of Species}]. While KBs have been a key component of artificial intelligence since the field's inception \citep{newell1956logic, newell1959} broad-coverage KBs are inevitably incomplete \citep{min2013distant}, despite efforts to automate their creation through text extraction \cite{angeli2015leveraging, NELL-aaai15}.


An alternative to extracting information to augment existing KBs relies on directly
answering queries using text corpora \cite{chen2017reading}.
Recently, \citet{dhingra2019differentiable} proposed DrKIT which answers questions based on a \textit{virtual KB} (VKB) constructed automatically from a text corpus.  The key idea behind DrKIT is to greatly simplify the construction of a KB by building a ``soft KB'', closely related to the original text, and compensate for the lack of structure in the VKB by employing more sophisticated neural methods to answer questions.  In DrKIT, each element of the VKB is composed of an entity mention and its surrounding context, encoded into a dense embedding representation. Given a query, a learned encoder projects that query into the same embedding space of the VKB mentions. The query vector is scored amongst all embedded mentions resulting in the retrieval of relevant mentions, analogous to the retrieval of a triple from a standard structured KB.  Other neural operations can be used to differentiably reason over the VKB, for instance by answering ``multihop'' questions that combine information from multiple embedded mentions.


A weakness of DrKIT is that constructing the VKB relied on distant supervision using existing KB triples. This leads to two limitations: (1) it is unclear if the VKB will correctly encode relations not present in the original structured KB used for distant supervision, and (2) the approach is inapplicable in domains without existing structured KBs (such as many technical areas.)  To address these problems, we introduce (1) a novel VKB construction method which can be trained without any distant KB supervision and (2) a set of differentiable reasoning operations on the VKB, which we call the Open Predicate Query Language (\tae{}). Hence, unlike DrKIT's procedure, an OPQL virtual KB can be created from any entity-linked corpus.



The key idea in constructing the \tae{} VKB is to use a dual-encoder pre-training process.  Our process is similar to that used in \cite{baldini-soares-etal-2019-matching}, which was shown to be useful for tasks such as relation classification.  Here we show that this pre-training process can also construct a VKB that effectively supports more complex reasoning operations, and also that \tae{} operations can be tightly integrated with a neural language model (LM).

To summarize this paper's contributions: (1) we describe an effective VKB pre-training method that, unlike previous methods, does not require an initial structured KB for distant supervision, and (2) demonstrate that this VKB can be used to answer multi-hop semi-structured queries effectively---in fact (3) \tae{} outperforms previous state-of-the-art VKB methods significantly, despite needing less supervision.  We also (4) 
demonstrate that \tae{} can be injected as an external memory into a neural LM to obtain a new state-of-the-art on a widely-studied QA task \cite{webqsp}.  Finally we (5) extend our VKB-injected LM with the ability to combine multiple \tae{} operations, and demonstrate this leads to a new state-of-the-art  on a QA task requiring compositional and conjunctive reasoning \cite{talmor18compwebq}, even outperforming much larger text-to-text models.

%% file: model.tex
\section{Model \label{sec:model}}

\begin{figure*}[h!]
  \centering
  \includegraphics[width=0.9\textwidth]{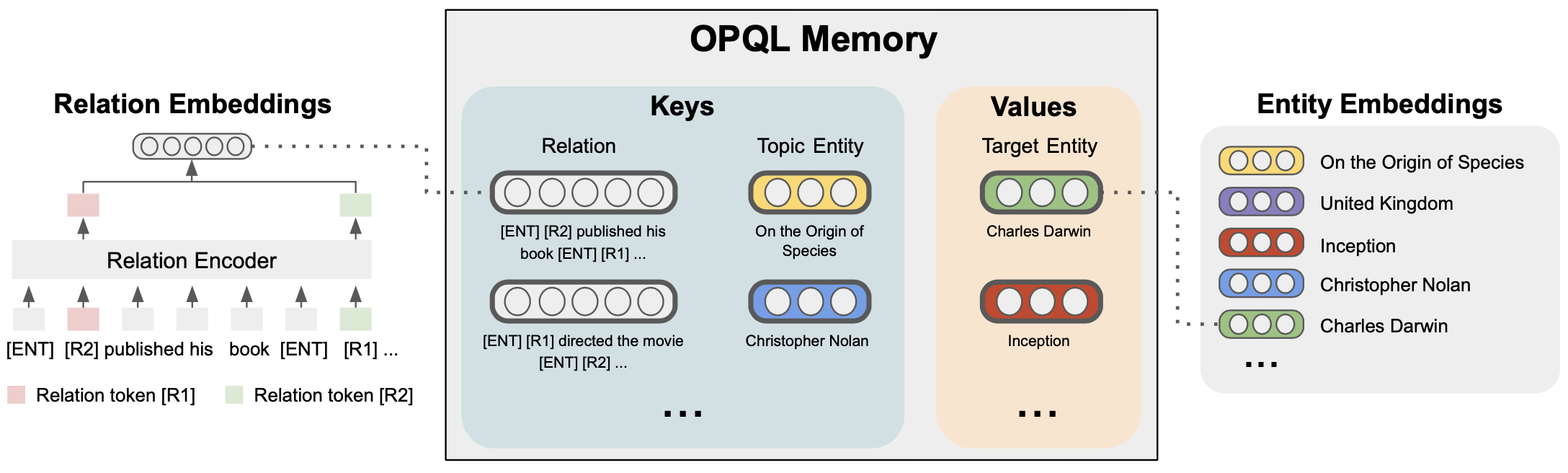}
  \vspace{-16pt}
  \caption{\textbf{\tae{} memory structure.} The \tae{} memory is a key-value memory. Keys are computed from embeddings of the topic entity, e.g. \textit{On the Origin of Species}, looked up from the entity embedding table, and relation embeddings from the pretrained relation encoder. Values are embeddings of target entities. The memory is constructed from any entity-linked text corpus, e.g.~Wikipedia.
  \label{fig:model}}
\end{figure*}

In this work, we propose \tae{}, a method for building a VKB from text without any structured supervision. \tae{} naturally supports compositional reasoning and can serve as the knowledge source for a memory augmented LM.
Entries in the \tae{} memory encode the relationship between pairs of entities, described in  natural language. For example, a sentence ``\textit{Charles Darwin} published his book \textit{On the Origin of Species} in 1859'' describes the authorship of entity \textit{On the Origin of Species}. 

Importantly, these unstructured relationships expressed in text can be extremely fine-grained, covering semantics that would never be included in a pre-defined KB schema. This is made possible because \tae{} has the ability of learning without any structured supervision.
These relationships are organized into a key-value memory index (\S \ref{sec:tae_memory}).
Each key is the composition of a topic entity (e.g. \textit{On the Origin of Species}) and an associated latent relationship expressed in text, constructed using pretrained entity embeddings (\S \ref{sec:entity-linking} and a pretrained text encoder (\S \ref{sec:rel-encoder}).
Its value is the corresponding target entity, in this case, \textit{Charles Darwin}.
When the memory is queried, the returned value can be used directly to answer the input query (\S \ref{sec:relation_following}) or it can be integrated into an LM for further reasoning before producing a final prediction (\S \ref{sec:memory_lm}). 

\subsection{Background} \label{sec:background}
\noindent\textbf{Input} The input to the model used for pretraining OPQL is a sequence of $n$ tokens $C=[c_0, c_1 ... c_n]$ for an arbitrary text span, e.g. ``\textit{Charles Darwin} published his book \textit{On the Origin of Species} in 1859'',
that contains a set of demarcated entity mentions $M$. A mention ``\textit{Charles Darwin}'' $ \in M$ is denoted as $m=(c_i, c_j, e_m) \in M$, where $i$ and $j$ denote the first and last token of the mention span and the mention is linked to entity $e_m$ in a predefined entity vocabulary $\mathcal{E}$, e.g. {Charles\_Darwin} (Q1035) in Wikidata. 

\noindent\textbf{Enity Pairs} \tae{} learns to encode the relationship between a pair of entities $(e_1, e_2)$, where $e_1$ is referred as topic entity and $e_2$ as target entity. In the example above, \textit{On the Origin of Species} is the topic entity $e_1$ and \textit{Charles Darwin} is the target entity $e2$. Potentially\footnote{Heuristics for limiting the number of entity pairs derived from text are discussed in Section~\ref{sec:pre-train}.} each distinct entity pair in a sentence can be encoded. The relationship between a pair of entities is directional, so the pairs $(e_1, e_2)$ and $(e_2, e_1)$ are encoded differently.

\noindent\textbf{Entity Embeddings} A global entity embedding table $\textbf{E} \in \mathbb{R}^{|\mathcal{E}| \times d_e}$ is pretrained using the RELIC strategy \cite{ling2020learning}; here $\mathcal{E}$ is the entity vocabulary and $d_e$ is the  embedding dimensionality.  Similar to pretraining a masked language model (MLM), an entity mention is randomly masked from the context and the goal is to retrieve the masked entity from the entity vocabulary $\mathcal{E}$.

\noindent\textbf{Key-Value Memory} We structure \tae{} as a key-value memory. A key-value memory (\textbf{K}, \textbf{V}) is a general way of storing and retrieving knowledge \cite{miller2016key}. When queried, key embeddings $\textbf{k}_i \in \textbf{K}$ are matched with the query vector, and the corresponded value embedding $\textbf{v}_i$ is returned.  Each entry in \tae{} encodes a pair of entities $(e_1, e_2)$, along with a piece of text $C$ that describes the relationship between $e_1$ and $e_2$. The key-value memory is constructed from the pretrained entity embedding table \textbf{E} and relation embeddings precomputed from a pretrained relation encoder. A key memory holds information from the topic entity and the relationship, and a value memory holds the target entity. We will discuss the pretrained entity and relation embeddings first and then discuss how to construct the key-value memory using the pretrained embeddings.





\subsection{Relation Encoder}\label{sec:rel-encoder}



\textbf{Preprocessing Text for the Relation Encoder}
\tae{} encodes the relationship between a pair of entities $(e_1, e_2)$ in a natural language sentence $C$ where both $e_1$ and $e_2$ are mentioned. Intuitively $C$ describes the relationship between $e_1$ and $e_2$ in the context, and we train a relation encoder to represent the relationship as a vector.

To indicate the location of the topic and target entities $e_1$ and $e_2$ we introduce two special tokens [R1] and [R2] that are inserted directly after the mentions of the topic and target entities in the sentence (e.g., $C_r$ becomes ``\textit{Charles Darwin} [R2] published his book \textit{On the Origin of Species} [R1] in 1859''). The contextual encodings of [R1] and [R2] will be used to compute relation embeddings.  We also introduce another special token [ENT] to mask the topic and target entity mentions; thus sentence $C_r$ finally becomes ``[ENT] [R2] published his book [ENT] [R1] in 1859''. Masking the mentions of entities prevents the relation encoder from memorizing the surface forms of the entities, and helps it generalize to similar relations involving other entities. The contextual embeddings at [ENT] tokens are not used by the relation encoder, but will be used for (masked) entity linking in \S \ref{sec:memory_lm}. 

\textbf{Computing Relation Embedding} Given this preprocessing of an entity-mention pair $(e_1, e_2)$, the relation embedding $\textbf{r}_{e_1, e_2}$ is defined as follows.
Let $s$ and $t$ be the location of tokens [R1] and [R2] in the masked sentence $C_r$. We construct a projection of the concatenation of the contextual embeddings $\textbf{h}_{s}$ and $\textbf{h}_{t}$ at the locations $s$ and $t$, as follows:
\begin{align} 
    \textbf{r}_{e_1, e_2} &= \textbf{W}_r^T ~ [\textbf{h}_{s}; \textbf{h}_{t}] \label{eq:relation_encoder}
\end{align}

\textbf{Training the Relation Encoder}
We train our relation encoder following  \citet{baldini-soares-etal-2019-matching}, who train embeddings such that relation mentions containing the same entity pairs are more similar to each other than relation mentions that contain different entity pairs. 

Specifically, mini-batches are constructed which contain at least two documents that contain entity pair $(e_1,e_2)$, as well as negative documents containing different relation pairs (see Section~\ref{sec:pre-train} for details.)  Use $\textbf{r}_{e_1, e_2}$ to denote the embedding from input $C$ that of $(e_1, e_2)$,  use $\textbf{r}_{e_i, e_j}$ for embeddings from documents $C'_0, \dots, C'_n$ of other pairs $(e_i, e_j)$'s, and let $\mathbb{I}_{e_i=e_1, e_j=e_2}$ be an equality indicator for entity pairs in the minibatch. We maximize the inner product between relation embeddings iff the same pair $(e_1, e_2)$ is mentioned in the candidates:
\begin{align}
L_{\textnormal{rel}} =
\textnormal{cross\_ent}(
\textnormal{softmax}({\textbf{r}_{e_1, e_2}}^T \textbf{r}_{e_i, e_j}), \mathbb{I}_{e_i=e_1, e_j=e_2}) \label{eq:rel-loss}
\end{align}

\subsection{Entity Linking} \label{sec:entity-linking}
Additionally, we use a multi-task training objective to learn representations of individual entities by learning to link other mentions in the context---i.e., mentions other than the topic and target entities---to the correct entity. For example, in the masked sentence $C_r$, ``[ENT] [R2], an \textit{English} naturalist, published his book [ENT] [R1] in 1859'', the mention \textit{English} is not part of the pair $(\textit{On the Origin of Species}, \textit{Charles Darwin})$ and thus not masked with [ENT], but should be linked to an entity for ``England''.  These \textit{context entity} mentions are represented as $m_{e_a} = (c_i, c_j, e_a)$ where $c_i$ and $c_j$ are the start and end positions, and $e_a$ is the entity to which this mention should be linked. 
We construct an embedding of mentions $m_{e_a}$ from the contextual embedding $\textbf{h}_i$ at the start position $c_i$ for the mention, i.e. $\textbf{m}_{e_a} = \textbf{W}_{e}^T \textbf{h}_i$.  This is used to retrieve the most similar entity from the embedding table $\textbf{E}$, scored with inner product distance, 
using this loss:
\begin{align}
L_{\textnormal{el}} = & \textnormal{cross\_ent} (\textnormal{softmax}({\textbf{m}_{e_a}}^T \textbf{e}_i), \mathbb{I}_{e_i=e_a}) \label{eq:el-loss}
\end{align}



\subsection{Storing \tae{}'s VKB as a key-value memory}\label{sec:tae_memory}
\tae{} stores relationships between pairs of entities in a key-value memory (\textbf{K}, \textbf{V}). 
Given an input $C$ that mentions a pair of entities $e_1$ and $e_2$,
the key embedding $\textbf{k}_{e_1, e_2}$ for the pair $(e_1, e_2)$
is constructed compositionally using the embeddings of topic entities $\textbf{e}_1$ from the entity embedding table $\textbf{E} \in \mathbb{R}^{|\mathcal{E}| \times d_e}$ and the pretrained relation embedding $\textbf{r}_{e_1, e_2}$. $\textbf{W}_k$ is a linear projection matrix that will be learned in the finetuning tasks. The value  $\textbf{v}_{e_1, e_2}$ is the embedding of the target entity $\textbf{e}_2$.
$$
\textbf{k}_{e_1, e_2} = \textbf{W}_k^T ~ [\textbf{e}_1; \textbf{r}_{e_1, e_2}] \in \textbf{K}, ~~~ \textbf{v}_{e_1, e_2} = \textbf{e}_2 \in \textbf{V}
$$


We iterate through all sentences in the Wikipedia corpus that contain two or more entities to construct the \tae{} memory. 
Note that each entity pair can be mentioned one or multiple times in the corpus. The memory could keep all mentions of an entity pair $(e_1, e_2)$, or instead reduce multiple mentions of the same entity pair to a single entry, for example averaging the key embedding of the individual mentions. This choice can be made based on application, or computational constraints. In the relational following task (\S \ref{sec:relation_following}), we keep all mentions of entity pairs in the \tae{} memory. For the open-domain QA task (\S \ref{sec:memory_lm}), the text corpus is larger,\footnote{We use the entire Wikipedia as our text corpus for the open-domain QA tasks.} and keeping all pairs of entities in the memory is not feasible. We randomly select up to 5 mentions for each entity pair and average their key embedding as the final entry in the \tae{} memory. (Note the the value embeddings of an entity pair are the same across multiple mentions).

\subsection{Pretraining Data} \label{sec:pre-train}
\textbf{Entity Linking Data} We use Wikipedia passages with hyperlinks as our pretraining. The hyperlinks link a span of tokens to a Wikipedia page for an entity $e$. The spans of tokens that are linked are considered as mentions $m$, and entity $e$ is the corresponding entity.
We take the top 1M entities that are most frequently mentioned in Wikipedia as our entity vocabulary $\mathcal{E}$. Passages are split into text pieces of 128 tokens. Entity linking is trained on the mentions of entities in the context, excluding the ones that are treated as topic and target entities. The pretraining corpus contains 93.5M mentions of the top 1M entities in the vocabulary.

\textbf{Relation Encoder Data} We first construct the vocabulary of entity pairs $(e_1, e_2)$ from the Wikipedia corpus. We count the pairs of entities that co-occur in the same text piece, and discard the that appears less than 5 times. The pairs are sorted by their point-wise mutual information (PMI).
The top 800k pairs of entities are selected as our candidates for training. During training, an input with pair $(e_1, e_2)$ will be paired with $2$ positive examples that mentions the same pair of entities, and $8$ hard negative examples $(e_i, e_j)$ that mention either the same topic or target entity, i.e. $e_i = e_1$ or $e_j = e_2$, but not both. Batch negatives are also included in pretraining. 30.6M training data on the top 800k entity pairs are constructed from the Wikipedia corpus.

\textbf{Pretraining Loss} The relation encoder is pretrained with entity linking loss $L_{\textnormal{el}}$ in Eq. \ref{eq:el-loss} and relation encoder loss $L_{\textnormal{rel}}$ in Eq. \ref{eq:rel-loss}, i.e. $L = L_{\textnormal{el}} + L_{\textnormal{rel}}$.

%% file: reasoning.tex
\section{Relation Following with \tae{}} \label{sec:relation_following}

In the previous section we introduced the method for pretraining the representations
in the VKB. Here we describe an application of the VKB to answering multi-hop questions.

\textbf{Background} A relation following operation $Y = X.\textnormal{follow}(R)$ maps a set of entities $X$ and a set of relations $R$ to a set of entities $Y$ such that $Y = \{ y ~|~ \exists x\in{}X, r\in{}R : r(x,y)\}$. One common application of the relation following operation is to solve QA tasks \cite{cohen2020scalable, sun2020faithful}. For example, a question ``Who is the author of \textit{On the Origin of Species}?'' can be answered by computing $X.\textnormal{follow}(R)$ with $X = \{\textit{On the Origin of Species}\}$ and $R = \{\textit{author\_of}\}$. In a learning task, $R$ is a weighted set of relations, and the model learns to predict the relation weights from the question.\footnote{The set of entities $X$ is the topic entities in the question, which in the experiments below is either provided, or easily obtained.}   The resulting relation following query is then executed with a key-value lookup into the OPQL memory. 

\textbf{Preprocessing Data for Fine-Tuning} We finetune the pretrained relation encoder to compute the queries for the relation following task as follows.
Let $Q$ be a question, which we will assume contains one known topic entity $e_1$. We let $X = \{e_1\}$ contain that entity. The model must predict a weighted set of target entities $Y$ that answer the question. Answers are always appended to the question, e.g. ``Who is the author of \textit{On the Origin of Species}? \textit{Charles Darwin}''. To encode the relation embeddings $\textbf{r}_{X, Y}$ for the preprocessed question/answer pair, [R1] and [R2] are inserted after the mentions of the topic entity X and target (answer) entity Y, and both mentions are masked with [ENT]. In the example above, the masked question thus becomes ``Who is the author of [ENT] [R1]? [ENT] [R2]''. This transformation ensures that the masked question has similar form to the input of the relation encoder, so the pretrained relation encoder can be easily finetuned for relation following tasks.

\textbf{Relation Following for OPQL} Recall that a follow query is executed with a key-value lookup from the OPQL memory.  Specifically, a query vector $\textbf{q}_{X, Y}$ is composed using the embedding of the topic entity $X$ and the relation embedding $\textbf{r}_{X, Y}$ of the question:
\begin{align} 
\textbf{e}_X &= \sum_i \alpha_i \cdot \textbf{e}_{x_i}, ~~ x_i \in X \\
\textbf{q}_{X, Y} &= \textbf{W}_q^T [\textbf{e}_X; \textbf{W}_t^T \textbf{r}_{X, Y}]  \label{eq:follow-query}
\end{align}
where in general $\textbf{e}_X$ is the weighted average of embeddings of entities $e_{x_i} \in X$ (this general case applies for multi-hop questions),  $\alpha_i$ is the weight of $x_i$ in $X$, and $\textbf{W}_t$ is a learned projection matrix.


The query $\textbf{q}_{X, Y}$ is then used to retrieve against the key memory $\textbf{K}$, returning the top $k$ entries with the largest inner product scores $(k_{e_i, e_j}, e_j)$.\footnote{The value $v_{e_i, e_j}$ of the entry is always $e_j$, so we write it as $(k_{e_i, e_j}, e_j)$ for short.} The $T_k$ set of top-k retrieved values $\{e_j\}$ are the answers $Y$. The weights $\beta_{{e_i, e_j}}$ of entity $e_j$ is the softmax of scores of the top $k$ retrieval result, denoted as $\textnormal{top}_k (\textbf{q}_{X, Y}, \textbf{K})$.
\begin{align*}
   \beta_{{e_i, e_j}} &= 
     \left\{ \begin{array}{l}
         \textnormal{softmax}(\textbf{q}_{X, Y}^T ~ \textbf{k}_{e_i, e_j}
            \mbox{~for~$(e_i, e_j) \in T_k$}) \\ 
         0, ~~~ \mbox{~else}
        \end{array}
     \right. 
\end{align*}

To improve retrieval accuracy, we also apply a sparse filter on the retrieval result to ensure the topic entity of the retrieved pair $(e_j, e_k)$ is in $X$ \citep{seo-etal-2019-real,dhingra2019differentiable}. To train the parameters of the relation
following operation we optimize the retrieved set of answers against the ground truth using the loss
$$
L_{\textnormal{follow}} = \textnormal{cross\_ent}(\beta_{{e_i, e_j}}, \mathbb{I}_{e_j \in \textnormal{Ans}})
$$

\textbf{Multi-hop Relational Following} Relation following operations can be chained to answer multi-hop questions: e.g., two-hop question can be decomposed into $Y = X.\textnormal{follow}(R_1).\textnormal{follow}(R_2)$. In this case, the predicted set of intermediate entities from the previous hop $X^{(t)} = X^{(t-1)}.\textnormal{follow}(R_{t-1})$ becomes the input to the next hop, i.e. $X^{(t+1)} = X^{(t)}.\textnormal{follow}(R_{t})$. In our model we use a single relation embedding $\textbf{r}_{X, Y}$ for the question but learn a different projection matrix $\textbf{W}_t^{(t+1)}$ for each hop: e.g.~we use
$$
\textbf{q}_{X, Y}^{(2)} = \textbf{W}_q^T [\textbf{e}_{X^{(1)}}; {\textbf{W}_t^{(2)}}^T \textbf{r}_{X, Y}]
$$
to form the query $\textbf{q}_{X, Y}^{(2)}$ for the second hop. 
In the multi-hop setting, the loss is only computed at the last step.
The number of hops is a hyper-parameter and in our experiments it is given for each dataset. 

\textbf{Finetuning Details} For the relational following task, the relation embeddings of entity pairs in the memory is precomputed and fixed at finetuning time. The pretrained relation encoder is finetuned to compute $\textbf{r}_{X, Y}$ in Eq.\ref{eq:follow-query} for the query vector $\textbf{q}_{X, Y}$. We also trained the projection matrices $\textbf{W}_q$ and $\textbf{W}_t^{(\cdot)}$, but fixed the entity embedding table $\textbf{E}$. The finetuning job is only trained with the loss $L_{\textnormal{follow}}$, since entity and relation embeddings are fixed in this task.




\subsection{Experiments: Reasoning Over VKB \label{sec:exp_reasoning}}
\subsubsection{Datasets}
\textbf{MetaQA} \cite{zhang2017variational} is a multi-hop QA dataset that extends the WikiMovies dataset to 2-hop and 3-hop questions in the movie domain. The questions are generated from templates, e.g. "Who starred in the movies directed by Christopher Nolan". Questions in MetaQA are answerable by the corpus in the original WikiMovies dataset. The corpus contains 18k passages that describes 7 different relations between 43k entities. We extracted 106k entity pairs from the corpus whose topic entity is the Wikipedia page title and target entity is one of the mentions in the passage. Reverse pairs are included, for a total of 213k entity pairs.

\textbf{Multi-hop Slot Filling (MSF)} \cite{dhingra2019differentiable} presents a large scale multi-hop reasoning dataset constructed from WikiData that contains 120k passages, 888 relations, and more than 200k entities. Queries are constructed from multi-hop paths in WikiData and turned into natural language questions by concatenating the head entity with a series of relations, e.g. (``Steve Jobs, founder, headquarter in, ?''). Similar to MetaQA, we constructed entity pairs by taking the Wikipedia page title as the topic entity and the entities mentioned in the passages as target entities.
We extracted 1.3m and 781k entity pairs for 2-hop and 3-hop questions respectively.

\subsubsection{Baselines}
GRAFT-Net \cite{sun2018open} is a GCN based model that can perform reasoning jointly over text and knowledge bases. PullNet \cite{sun2019pullnet} extends GRAFT-NET by introducing an iterative retrieve-and-classify mechanism to solve multi-hop questions. PIQA \cite{seo-etal-2018-phrase} and DrKIT \cite{dhingra2019differentiable} build mention-level indexes using a pretrained encoder. 
Training the DrKIT index requires distant supervision using artificial queries constructed from KB. KV-Mem \cite{miller2016key} and DrQA \cite{chen2017reading} are another two widely-used open-domain QA baselines.

\subsubsection{Results}

We experiment on the relational following task on both a pretrained memory (\tae{}-pretrained) and a finetuned memory (\tae{}). The Hits@1 results of the model are listed in Table \ref{tab:reasoning}. The \tae{}-pretrained memory directly applies the pretrained relation encoder on entity pairs extracted from the dataset corpus, without any finetuning of the relation encoder. 
\tae{}-pretrained achieves the state-of-the-art performance on both MSF 2-hop and 3-hop datasets, though \tae{}-pretrained is slightly lower than DrKIT on the MetaQA. This is because DrKIT finetuned its mention encoder on MetaQA corpus. The MetaQA corpus only contains 14 relations (including their inverse) in the movie domain, so it's easy for the model to learn only capturing these relationship between entities. To mitigate this bias, we finetune the \tae{} relation encoder on MetaQA corpus. The finetuning data is distantly constructed from 1-hop questions in the MetaQA dataset, by masking the topic entity from the question and inserting a placeholder for the target entity at the end of the question. We end up with 10K finetuning data for MetaQA and 19K for MSF. \tae{} with finetuned memory outperforms DrKIT on 2-hop questions by 2.5 points and is very comparable on 3-hop questions.

\begin{table}[t]
    \small
    \centering
\begin{tabular}{l|cc|cc}
 \toprule
  & \multicolumn{2}{c}{MetaQA} & \multicolumn{2}{|c}{MSF} \\
  Model & 2Hop & 3Hop & 2Hop & 3Hop \\
  \midrule 
  KVMem  & 7.0 & 19.5 & 3.4  & 2.6 \\
  DrQA  & 32.5 & 19.7 & 14.1 & 7.0 \\
  GRAFT-Net  & 36.2 & 40.2 & -  & - \\
  PullNet  & 81.0 & 78.2 &  - & - \\
  PIQA  & -  & - & 36.9 & 18.2  \\
  DrKIT & 86.0 & \textbf{87.6}  & 46.9 & 24.4 \\
  \midrule
  \tae{}-pretrained  & 84.7 & 84.3  &48.5 & 28.1\\
  \tae{} & \textbf{88.5} & 87.1  & \textbf{49.2} & \textbf{29.7}  \\
\bottomrule
\end{tabular}
  \vspace{-12pt}
    \caption{Hits@1 results on multi-hop relational following task.\label{tab:reasoning}}
\end{table}


\subsubsection{Generalization to Novel Relations}
\begin{figure}[h!]
    \centering
    \includegraphics[width=0.8\textwidth]{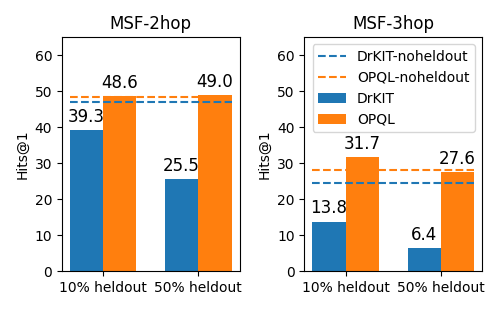}
    \vspace{-12pt}
    \caption{Hits@1 on multi-hop queries containing at least one novel relation. The dashed lines represent the accuracy on the full dataset from Table \ref{tab:reasoning}.}
    \label{fig:abl-holdout-rel}
\end{figure}

The previous state-of-the-art model, DrKIT, uses training data in its
pretraining procedure that is distantly constructed from KB. This restricts the capacity of DrKIT to only encode KB relations observed at pretraining time.
However, pretraining \tae{} 
does not require any signal from knowledge bases, so it can also encode relations that are out of the relation vocabulary in KB. 
To demonstrate this difference, we hold out some portions of relations in the pretraining phase of DrKIT but evaluate on queries where at least one of the held-out relations is required to answer the queries. 
This simulates a scenario where KBs are incomplete and limited in their relation vocabularies. 
We compare DrKIT to the pretrained \tae{} memory. The result is shown in Figure \ref{fig:abl-holdout-rel}. The performance of DrKIT drops significantly when evaluated on queries with novel relations not seen at pretraining time, while the performance of \tae{} is consistent.



%% file: memlm.tex
\section{Augmenting \tae{} with a LM} \label{sec:memory_lm}




Recent advances in LM architectures have incorporated large external memories which can lead to better performance given fewer activated parameters (retrieved memories are only sparsely activated, rather than dense models which utilize nearly every parameter on every input). \citet{verga2020facts} proposed Facts-as-Experts (FaE) which injected an external fact memory into an LM, increasing its performance on open-domain QA. However, this approach relied on a structured KB, suffering from many of the shortcomings discussed in previous sections such as limited coverage and applicability. In this section, we address these issues by replacing the KB-based fact memory of FaE with our \tae{} memory. 

\begin{figure}[h!]
  \centering
  \includegraphics[width=0.9\textwidth]{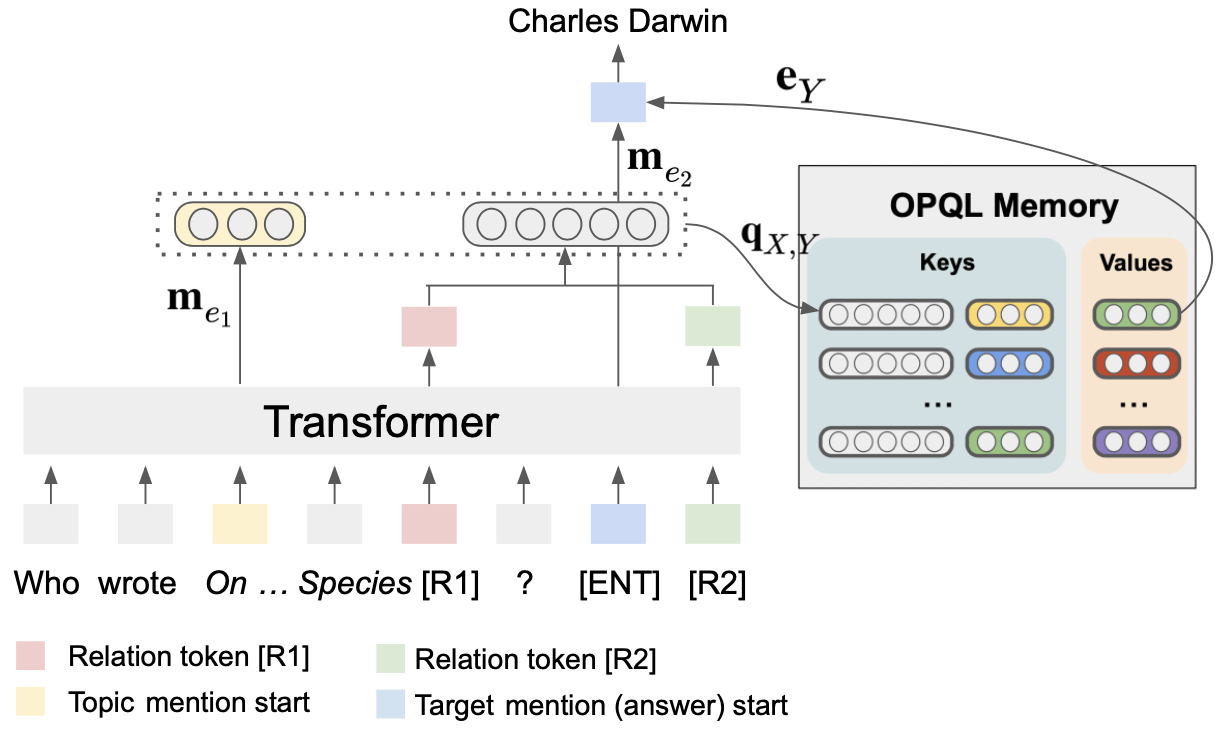}
  \vspace{-12pt}
  \caption{
  \label{fig:tae-lm} \textbf{\tae{}-LM model architecture}. A query vector $\textbf{q}_{X, Y}$ is constructed from the contextual embedding of the topic mention $\textbf{m}_{e_1}$ and the relation embedding $\textbf{r}_{X, Y}$, and retrieves the top few entries from the \tae{} memory. The retrieval results are aggregated into a single vector $\textbf{e}_Y$, and mixed with the contextual embedding of the masked mention (answer) $\textbf{m}_{e_2}$ to make the final prediction.}
\end{figure}

\textbf{Background} A memory injected language model, e.g. FaE \cite{verga2020facts}, learns to retrieve relevant entries from an external memory and mix the retrieved information into the language model to make its final predictions (see Figure \ref{fig:tae-lm}). \tae{} follows the same implementation as FaE but utilizes the pre-computed \tae{} memory as the external memory. The relation embedding in the \tae{} memory is both more diverse and fixed during training, leading to retrieval from the \tae{} memory being harder than the fact-based memory from FaE. To address this, we make several key changes to the original FaE architecture. First, the query vector is constructed as a function of the topic mention embedding and relation embedding. Second, we modify the learned a mixing weight between the memory retrieval results and language model predictions. Third, we extend \tae{}-LM to answer multi-hop questions. We will elaborate these changes in the rest of this section.\footnote{Please refer to the appendix and the \citet{verga2020facts} for more details.}

\textbf{Input} Similar to \S \ref{sec:relation_following}, the special tokens [R1] and [R2] are inserted directly after the mentions of topic entity $e_1$ and target entity (answer) $e_2$. But the mention of topic entity $e_1$ will not be masked with [ENT],
e.g. ``Who published the book \textit{On the Origin of Species} [R1] in 1859? [ENT] [R2]''. We denote the mention of topic entity as $m_{e_1}$. The mention $m_{e_1}$ is used to extract contextual mention embedding $\textbf{m}_{e_1}$ for the topic entity $e_1$, which will be used for entity linking (\S \ref{sec:entity-linking}), as well as constructing the query to retrieve from the \tae{} memory.
The training objective of \tae{}-LM is to predict the masked target entity (answer) $e_2$.

\textbf{Query Embedding} 
The query $\textbf{q}_{X, Y}$ is computed compositinoally by concatenating the contextual mention embedding $\textbf{m}_{e_1}$ of the topic mention $m_{e_1}$ (as described by entity linking \S \ref{sec:entity-linking}) and the relation embedding $\textbf{r}_{X, Y}$. Ideally, $\textbf{m}_{e_1}$ should be close to the embedding of the oracle topic entity $\textbf{e}_1$, which is supervised with entity linking loss (Eq. \ref{eq:el-loss}).
The relation embedding $\textbf{r}_{X, Y}$ comes from the relation encoder (Eq. \ref{eq:relation_encoder}) that operates on special tokens [R1] and [R2].\footnote{Since the mention of entity $e_1$ is not masked from the input, the relation embedding $\textbf{r}_{X, Y}$ may contain some leaked information from the topic entity $e_1$. One could encode the relation embedding $\textbf{r}_{X, Y}$ separately from an input where both $e_1$ and $e_2$ are masked. We do not take this solution as it doubles the computation cost of the expensive Transformer layers.}
$$
\textbf{q}_{X, Y} = \textbf{W}_q^T [\textbf{m}_{e_1}; \textbf{W}_t^T \textbf{r}_{X, Y}]
$$

\textbf{Mixing with LM} 
Let $\textbf{e}_Y$ be the retrieved embedding returned from the \tae{} memory and $\textbf{m}_{e_2}$ be contextual embedding of the masked mention [ENT] predicted from the language model. $\textbf{e}_Y$ and $\textbf{m}_{e_2}$ are mixed with a mixing factor $\lambda$. $\lambda$ decides whether retrieved embedding should be included to make final predictions. $\lambda$ should be large if there is a relevant entry with pair $(e_1, e_2)$ in the memory, so retrieving this pair can help predict the masked entity $e_2$. 
Different from \citet{verga2020facts}\footnote{FaE \cite{verga2020facts} introduced a \texttt{null} fact whose embedding $\textbf{k}_{\texttt{null}}$ is a learned variable.}, \tae{}-LM introduced a \texttt{null} relation to the \tae{} memory, whose embedding $\textbf{r}_{\texttt{null}}$ is a learned variable, and constructed a \texttt{null} entry with key embedding $\textbf{k}_{\texttt{null}} = \textbf{W}_k^T [\textbf{m}_{e_1}; \textbf{r}_{\texttt{null}}]$ and value embedding $\textbf{v}_{\texttt{null}} = \overrightarrow{\textbf{0}}$. The query is encouraged to retrieve the \texttt{null} entry if no relevant pair of $(e_1, e_2)$ exists in the \tae{} memory. We re-use the memory key projection matrix $\textbf{W}_k$ to compute the key embedding $\textbf{k}_{\texttt{null}}$.
\begin{align}
\textbf{m}'_{e_2} = \textbf{m}_{e_2} + (1 -\lambda) \cdot \textbf{e}_Y,  ~~ \lambda = \textnormal{softmax} (\textbf{q}_{X, Y}^T ~ \textbf{k}_{\texttt{null}}) \label{eq:mel-query}
\end{align}

\textbf{Multi-hop \tae{}-LM} Some open-domain questions require multi-hop reasoning to find the answers, e.g. ``Where is the author of \textit{On the Origin of Species} educated?''. To answer such questions, the model should first find the answer of the first-hop of the question, and use the it as topic entity to answer the second-hop of the questions. \tae{}-LM can naturally solve this problem by repeating the retrieval and mixing steps. 

Let ${\textbf{m}'_{e_2}}^{(t)}$ from Eq. \ref{eq:mel-query} be the memory-injected contextual embedding that was used to predict the answer of the $t$'th hop.
In the second hop, 
${\textbf{m}'_{e_2}}^{(t)}$ becomes the embedding of the topic entities and used to compute query $\textbf{q}_{X, Y}^{(t+1)}$ for the $(t+1)$'th hop. Again, we keep the relation embedding $\textbf{r}_{X, Y}$ unchanged, but learn another projection matrix $\textbf{W}_t^{(t+1)}$. ${\textbf{m}'_{e_2}}^{(t+1)}$ will be computed accordingly with updated $\textbf{m}_{e_2}^{(t+1)}$, $\lambda^{(t+1)}$ and $\textbf{e}_Y^{(t+1)}$ at the $(t+1)$'th hop.
\begin{align*}
\textbf{q}_{X, Y}^{(t+1)} &= \textbf{W}_q^T [{\textbf{m}'_{e_2}}^{(t)}; \textbf{W}_t^{(t+1)T} \textbf{r}_{X, Y}] 
\end{align*}


\subsection{Pretraining \tae{}-LM}
We propose a second stage pretraining for \tae{}-LM that learns 
to retrieve from the \tae{} memory and compute the mixing weight $\lambda$.
Relation embeddings $\textbf{r}_{e_1, e_2}$ are precomputed and fixed at the second stage pretraining. We continue training Transformer layers and the entity embedding table $\textbf{E}$ for entity linking. The second stage pretraining is not required for the relational following task (\S \ref{sec:relation_following}).

\textbf{Data} We use Wikipedia passages as our pretraining data. Given an input $C$ that contains $M$ mentions, we randomly select one mention as our target entity $e_2$ and mask it with [ENT]. All other mentions are considered topic entities $\{e_1\}$. Special tokens [R1] and [R2] are inserted after mentions of topic and target entities. Mentions of the topic entities will not be masked. Each entity pair $(e_1, e_2)$ will be treated independently.
In an example with three mentions, ``[ENT] [R2], an \textit{English} [R1] naturalist, published his book \textit{On the Origin of Species} [R1] in 1859.'', the mention $\textit{Charles Darwin}$ is selected as the target entity $e_2$. Retrieval and mixing steps are performed on both topic entities $\textit{English}$ and $\textit{On the Origin of Species}$ independently to predict the masked entity $\textit{Charles Darwin}$. \tae{}-LM is trained on 85.6M text pieces with a length of 128 tokens.

Please refer to the paper by \citet{verga2020facts} for more discussion on the loss terms and finetuning details. Besides the parameters in \citet{verga2020facts}, we additionally train the null relation embedding $\textbf{r}_{\textnormal{null}}$ and finetune the relation projection matrix $\textbf{W}_t^{(\cdot)}$.

\subsection{Experiments: Integrating VKB with LMs \label{sec:exp_lm}}
\vspace{-3pt}


Next, we experiment with \tae{} in another realistic scenario -- open-domain QA. In the relational following task (\S \ref{sec:relation_following}) where questions are often semi-structured and oracle topic entities are provided. In open-domain QA, questions are more diverse natural language and do not contain oracle-linked entities. In our experiments, we make a weaker assumption that mention detection has been run on the input providing boundaries of entity mentions to the model. \tae{} can effectively solve open-domain QA by retrieving relevant entries from the memory and return the corresponding values as answers and the retrieval results from the memory can be mixed with language model predictions to further improve the performance. We call this \tae{}-LM.

\vspace{-3pt}
\subsubsection{Dataset}
\vspace{-3pt}
\textbf{WebQuestionSP} (WebQSP) \cite{webqsp} is an open-domain Question Answering dataset that contains 4737 factual questions posed in natural languages. Answers to the questions are labeled with entities in Freebase. Since the pretraining was performed on Wikidata, we convert the Freebase entity ids (MIDs) to Wikidata ids (QIDs). After the conversion, 88.2\% questions are answerable by QIDs. 

\textbf{ComplexWebQuestions} (ComplexWebQ) \cite{talmor18compwebq} extends WebQuestionsSP to multi-hop questions. The ComplexWebQuestions dataset contains 34,689 complex questions, including 45\% composition questions, 45\% conjunction questions\footnote{See Appendix for details of solving conjunction questions.}, and 10\% others. Similar to the WebQuestionsSP dataset, we convert Freebase MIDs to Wikidata QIDs. 94.2\% of the questions are answerable.

\vspace{-3pt}
\subsubsection{Baselines}
\vspace{-3pt}
We compare \tae{} with several strong open-domain QA baselines. GRAFT-Net and PullNet \cite{sun2018open, sun2019pullnet} are Graph-CNN based models that are introduced in the previous experiments. BART \cite{lewis2019bart} and T5 \cite{raffel2019exploring} are large pretrained text-to-text Transformer models.\footnote{WebQuestionsSP is finetuned on T5-11B \cite{verga2020facts}. Due to hardware constraint, we finetune T5-3B for ComplexWebQuestions.} EaE \cite{fevry2020entities} is trained to predict masked entities, but without an external (virtual) KB memory. 
EaE does not have an external reasoning module, so it requires the reasoning process performed all from the Language Model. 
DPR \cite{karpukhin2020dense} is a retrieve-and-read approach that pairs a dense retriever in the embedding space with a BERT based reader. To extend DPR to handle multi-hop questions, we concatenate answers from the previous hops to the question as the query for the next hop. This is referred as DPR-cascade.\footnote{We run DPR and DPR-cascade with the pretrained checkpoint on Natural Questions \cite{nq}. Finetuned DPR reader on ComplexWebQuestions only gets 18.2\% Hits@1.} We also include results of FaE though its external memory is built from KB.

\subsubsection{Results}
\tae{}-LM outperforms the baseline models on both datasets (Table \ref{tab:lmqa}). We also experimented with an ablated model \tae{}-follow that only performs multi-hop relational following on questions,\footnote{We do not assume oracle entity linking here. The model use the contextual embedding of the topic entity to construct the query.} i.e. retrieved embeddings $\textbf{e}_Y$ is directly used
to make predictions. In WebQuestionsSP, the accuracy of retrieving the relevant pair is 85.4\% and that leads to 46.6\% accuracy of the WebQuestionsSP dataset. However, in ComplexWebQuestions, the coverage of \tae{} memory to answer both hops of the questions is only 22.1\% for compositional questions.\footnote{This number is computed from the intermediate answers provided in the dataset. We do not use the intermediate answers in training.} So the accuracy of \tae{}-follow is bounded. Mixing \tae{} with LM can further improve the performance of \tae{}-follow by 5.3\% on WebQuestionsSP and 22.4\% on ComplexWebQuestions. 

The entity-dependent null embedding $\textbf{k}_{\textnormal{null}}$ is crucial for \tae{}-LM. In an ablation study that the null embedding does not depend on the topic entity $e_1$, i.e. $\textbf{k}_{\textnormal{null}}$ is a learned variable shared by all entities, the performance on WebQuestions drops from 51.9 to 46.3 with the retrieval accuracy dropping from 85.4 to 69.5. The accuracy on Complex WebQuestions drops from 40.7 to 19.3.
\begin{table}[h]
    \small
    \centering
\begin{tabular}{l|ccc}
 \toprule
  Model & WebQSP & ComplexWebQ (dev) \\
  \midrule
  GRAFT-NET  & 25.3  & 10.6 \\
  PullNet  & 24.8 & 13.1 \\
  BART-Large  & 30.4  & - \\
  EaE  & 47.4  & 31.3 \\
  DPR  & 48.6  & 24.6 \\
  DPR-cascade & - & 25.1 \\
  T5  & 49.7  & 38.7 \\
  \midrule
  \tae{}-follow & 46.6 & 18.5 \\
  \tae{}-LM  & \textbf{51.9} & \textbf{40.7}\\
  ~~ \textit{+ webqsp pairs} & \textit{53.7}  & \textit{41.1} \\

  \midrule
  \midrule
  FaE  & 54.7 & -\\
  SoA (KB) & 69.0 (F1)  & 47.2 \\
\bottomrule
\end{tabular}
\vspace{-12pt}
\caption{\label{tab:lmqa} Hits@1 performance on open-domain QA datasets. \tae{}+webqsp pairs injects additional data specific memories. The state-of-the-art model on WebQSP (NSM \cite{liang2017neural}) and ComplexWebQ (PullNet-KB \cite{sun2019pullnet}) both use Freebase to answer the questions. FaE \cite{verga2020facts} has an external memory with KB triples.
}
\end{table}

\subsubsection{Injecting Entries into the \tae{} Memory}
We show that \tae{} can efficiently injecting new pairs to the memory to improve the coverage of knowledge at finetuning time, without having to retrain the relation encoder or LM. In the WebQuestionsSP experiment above, the pretrained \tae{} memory that contains 1.6M popular entity pairs (800k pairs plus their inverse) only covers 54.6\% of the questions, and each question entity was only included in 8.4 pairs in the memory of pre-selected entity pairs. To improve the coverage, we added 100 pairs per question entity with the highest PMI to the memory. The updated memory contains 1.8M pairs and the coverage increases to 82.9\%.

The result 
with the updated memory is presented in Table \ref{tab:lmqa}. 
The retrieval accuracy on questions that has relevant pairs in the memory drops from 85.4\% to 62.2\% as a result of adding $10$ times more pairs for each question entities, but the overall Hits@1 accuracy of the model improves from 51.9\% to 53.7\%. We also finetune \tae{}-LM on ComplexWebQuestions dataset. The improvement is less significant since retrieval is harder on complex questions.

%% file: related_work.tex
\section{Related Work \label{sec:related_work}}
Automatically extracting triples from a text corpus has been pursued for many years as a means of improving the coverage of KBs \citep{NELL-aaai15}.  Rather than extracting triples into a predefined vocabulary, OpenIE (typically) uses linguistic patterns to extract open vocabulary relations from text \cite{etzioni2008open, fader2011identifying}. \citet{gupta-etal-2019-care}) build an Open Knowledge Graph over Open IE extractions similar to a VKB. 

A few methods \citep{seo-etal-2018-phrase, dhingra2019differentiable} have built pre-computed memories of mention embeddings which are used directly to answer questions.  
These methods are most similar to our work though \tae{} differs by embedding mention pairs, rather than single mentions, and additionally, our method requires no structured supervision.
Another related of work proposed to build a passage level index \cite{karpukhin2020dense} to improve retrieval accuracy in the ``retrieve and read'' pipeline. Multi-hop retrieval models \cite{qi2020retrieve} are proposed for complex questions. 
Other approaches propose to retrieve embedded passages which are then passed to an LM to reason over  \cite{karpukhin2020dense, lewis2020retrieval, guu2020realm, lee2019latent}. In contrast \tae{} does not include a separate model for reading retrieved documents. 

Other work has shown that injecting an external memory constructed from KB into a LM can help training LMs \cite{peters2019knowledge}, improve generation tasks, \cite{logan2019barack}, and enable reasoning over an updated memory \cite{verga2020facts}. 
Additionally, our model's memory scales to millions of entries, whereas most prior systems that use KB triples have been with only a few hundreds of triples in the model at any point, necessitating a separate heuristic process to retrieve candidate KB triples \citep{ahn2016neural, henaff2016tracking, weissenborn2017dynamic, chen2018neural, mihaylov-frank-2018-knowledgeable, Logan_2019}.


















%% file: appendix.tex
\section{Appendix}
\subsection{\tae{}-ML Retrieval and Mixing Details}
\textbf{Retrieval from \tae{} Memory} Retrieving from the \tae{} memory is analogous to running a relational following operation $Y = X.\textnormal{follow}(R)$ with a learned set of relations $R$, but the set of topic entities $X$ is also unknown. 
Recall that the relational following task uses the weighted $X$ to compute its query embedding $\textbf{q}_{X, Y} = \textbf{W}_q^T [\textbf{e}_{X}; \textbf{W}_t^T \textbf{r}_{X, Y}]$. In the open-domain QA task, we do not assume oracle entity linking is provided, and thus not able to compute $\textbf{e}_X$ explicitly from set $X$. Instead, we consider the contextual mention embedding $\textbf{m}_{e_1}$ as an approximation of the centroid $\textbf{e}_X$. Ideally, $\textbf{m}_{e_1}$ should be close to the embedding $\textbf{e}_1$ of the oracle topic entity.
The query $\textbf{q}_{X, Y}$ is computed compositinoally by concatenating the contextual mention embedding $\textbf{m}_{e_1}$ and the relation embedding $\textbf{r}_{X, Y}$. The relation embedding $\textbf{r}_{X, Y}$ comes from the relation encoder (Eq. \ref{eq:relation_encoder}) that operates on special tokens [R1] and [R2]. 
$$
\textbf{q}_{X, Y} = \textbf{W}_q^T [\textbf{m}_{e_1}; \textbf{W}_t^T \textbf{r}_{X, Y}]
$$

Since the mention of entity $e_1$ is not masked from the input, the relation embedding $\textbf{r}_{X, Y}$ may contains some leaked information from the topic entity $e_1$. One potential fix is to encode the relation embedding $\textbf{r}_{X, Y}$ separately from an input where both $e_1$ and $e_2$ are masked. We do not take this solution as it doubles the computation cost of the expensive Transformer layers.






As discussed in \S \ref{sec:relation_following}, the result of the relational operation $Y = X.\textnormal{follow}(R)$ contains values $\{e_j\}$ of the top $k$ retrieved pairs $\{(e_i, e_j)\}$, s.t. $(e_i, e_j) \in \textnormal{top}_k(\textbf{q}_{X, Y}, \textbf{K})$. The weight $\beta_{e_i, e_j}$ is the softmax of the retrieval score.
We aggregate the embeddings of the retrieved entities $e_j \in Y$ into a single vector $\textbf{e}_Y$. $\textbf{e}_Y$ will be mixed with the contextual embedding of the masked mention [ENT] to predict the masked target entity $e_2$. 
\begin{align}
\textbf{e}_Y &= \sum_{(e_i, e_j)} \beta_{e_i, e_j} \textbf{e}_j, ~~(e_i, e_j) \in \textnormal{top}_k(\textbf{q}_{X, Y}, \textbf{K}) \label{eq:mel-retrieval} \\
L_\textnormal{re} &= \textnormal{cross\_entropy}(\beta_{{e_i, e_j}}, \mathbb{I}_{e_j \in \textnormal{Ans}}) \label{eq:retrieval-loss}
\end{align}

\textbf{Mixing with LM} The language model computes the contextual embedding of the masked entity $e_2$. Let $m_{e_2} = (c_i, c_i, e_2)$ be the masked mention [ENT] of the target entity (answer) $e_2$ that locates at the $i$'th token of the input sequence. The contextual embedding $\textbf{m}_{e_2}$ is a projection of Transformer output $\textbf{h}_i$ at the token $c_i$, that shares the same projection matrix $\textbf{W}_e$ with the entity linking task in \S \ref{sec:entity-linking}.
\begin{align}
    \textbf{m}_{e_2} = \textbf{W}_e^T \textbf{h}_i \label{eq:mel-lm}
\end{align}

The retrieved embeddings $\textbf{e}_Y$ (Eq. \ref{eq:mel-retrieval}) and contextual embeddings $\textbf{m}_{e_2}$ (Eq.\ref{eq:mel-lm}) of the masked mention are mixed with a mixing factor $\lambda$. $\lambda$ decides whether retrieved memory should be added to the contextual embedding. $\lambda$ should be large if there is a relevant entry with pair $(e_1, e_2)$ in the memory, so retrieving this pair can help predict the masked entity $e_2$. As suggested by \citet{verga2020facts}, we introduced a \texttt{null} relation to the \tae{} memory, whose embedding $\textbf{r}_{\texttt{null}}$ is a learned variable, and constructed a \texttt{null} entry with key embedding $\textbf{k}_{\texttt{null}} = \textbf{W}_k^T [\textbf{m}_{e_1}; \textbf{r}_{\texttt{null}}]$ and value embedding $\textbf{v}_{\texttt{null}} = \overrightarrow{0}$. The query is encouraged to retrieve the \texttt{null} pair if no relevant pair of $(e_1, e_2)$ exists in the \tae{} memory. We re-use the projection matrix $\textbf{W}_k$ to compute the key embedding $\textbf{k}_{\texttt{null}}$.
\begin{align}
\lambda &= \textnormal{softmax} (\textbf{q}_{X, Y}^T ~ \textbf{k}_{\texttt{null}}) \\
\textbf{m}'_{e_2} &= \textbf{m}_{e_2} + \lambda \cdot \textbf{e}_Y 
\end{align}

The memory-injected contextual embedding $\textbf{m}'_{e_2}$ is used to predict the masked entities $e_2$. 
\begin{align}
    L_\textnormal{mel} &= \textnormal{cross\_entropy}(\textnormal{softmax}({\textbf{m}'_{e_2}}^T ~ \textbf{e}_i), \mathbb{I}_{e_i \in \textnormal{Ans}}) \label{eq:mel-loss}
\end{align}

\textbf{Loss} Besides the entity linking loss on masked mention $\textbf{m}'_{e_2}$ (Eq. \ref{eq:mel-loss}), we jointly train entity linking on topic entity $e_1$ (Eq. \ref{eq:el-loss}), and also provide supervision on \tae{} memory retrieval (Eq. \ref{eq:retrieval-loss}).
$$
L_\textnormal{\tae{}-LM} = L_\textnormal{el} + L_\textnormal{re} + L_\textnormal{mel}
$$

\subsection{\tae{}-LM for Conjunction Questions}
A conjunction questions, e.g. ``Which \textit{English} author publish his book \textit{On the Origin of Species}?'', contains more than one topic entity \textit{English} and \textit{On the Origin of Species}. Both topic entities can potentially help to predict the answer, e.g. with pairs (\textit{English}, \textit{Charles Darwin}) that describes his nationality and (\textit{On the Origin of Species}, \textit{Charles Darwin}) that describes his publications.
We convert the input by appending the masked answer to the end of the question and inserting the special tokens [R1] and [R2] accordingly, e.g. ``Which \textit{English} [R1] author publish his book \textit{On the Origin of Species} [R1]? [ENT] [R2]''. 

A query vector $\textbf{q}_{X, Y, e_i}$ is constructed for each topic entity $e_i$, with the retrieval results $\textbf{e}_{Y, e_i}$ returned from the memory. All retrieved embeddings are mixed with the contextual embedding $\textbf{m}_{e_2}$ to predict the final answer
$$
\textbf{m}'_{e_2} = \textbf{m}_{e_2} + \sum_{e_i} \lambda_i \cdot \textbf{e}_{Y, e_i}
$$
where $\lambda_i$ is the mixing weight that is determined by each query independently.